\documentclass[runningheads]{llncs}
\usepackage{graphicx}

\usepackage{tikz}
\usepackage{booktabs}
\usepackage{comment}
\usepackage{amsmath,amssymb} % define this before the line numbering.
\usepackage{color}
\usepackage{subfig}
\usepackage[accsupp]{axessibility}  % Improves PDF readability for those with disabilities.

\usepackage{ruler}
\usepackage[width=122mm,left=12mm,paperwidth=146mm,height=193mm,top=12mm,paperheight=217mm]{geometry}
\newif\ifdraft
\draftfalse
\drafttrue

\ifdraft

\else
 \newcommand{\PF}[1]{}
 
 \newcommand{\KY}[1]{}
 
 \newcommand{\MS}[1]{}
 
 \newcommand{\ZD}[1]{}
 
 \newcommand{\YH}[1]{}
 
 \newcommand{\SPE}[1]{}
 
 \newcommand{\WJ}[1]{}
 
\fi

\begin{document}
% \linenumbers
\pagestyle{headings}
\mainmatter
\def\ECCVSubNumber{xxx}  % Insert your submission number here
\title{Unsupervised Learning of 3D Semantic Keypoints with Mutual Reconstruction } % Replace with your title

\begin{comment}
\titlerunning{ECCV-22 submission ID \ECCVSubNumber} 
\authorrunning{ECCV-22 submission ID \ECCVSubNumber} 
\author{Anonymous ECCV submission}
\institute{Paper ID \ECCVSubNumber}
\end{comment}
%******************

% \begin{comment}
\titlerunning{Mutual Reconstruction}
\author{Haocheng Yuan\inst{1}, Chen Zhao\inst{2}, Shichao Fan\inst{1}, Jiaxi Jiang\inst{1} and
Jiaqi Yang*\inst{1}}
\institute{Northwestern Polytechnical University, China\\
\email{\{hcyuan, fsc\_smile, jshmjjx\}@mail.nwpu.edu.cn},\\
\email{jqyang@nwpu.edu.cn},
\and
École Polytechnique Fédérale de LausanneL, Switzerland\\
\email{chen.zhao@epfl.ch}}
% \end{comment}
%******************
\maketitle

\begin{abstract}
Semantic 3D keypoints are category-level semantic consistent points on 3D objects. Detecting 3D semantic keypoints is a foundation for a number of 3D vision tasks but remains challenging, due to the ambiguity of semantic information, especially when the objects are represented by unordered 3D point clouds. Existing unsupervised methods tend to generate category-level keypoints in implicit manners, making it difficult to extract high-level information, such as semantic labels and topology. From a novel mutual reconstruction perspective, we present an unsupervised method to generate consistent semantic keypoints from point clouds explicitly. To achieve this, the proposed model predicts keypoints that not only reconstruct the object itself but also reconstruct other instances in the same category.
To the best of our knowledge, the proposed method is the first to mine 3D semantic consistent keypoints from a mutual reconstruction view. Experiments under various evaluation metrics as well as comparisons with the state-of-the-arts demonstrate the efficacy of our new solution to mining semantic consistent keypoints with mutual reconstruction.
\begin{figure}[h]
    \centering
    \subfloat[Previous Methods]{
    \includegraphics[width = .25\linewidth]{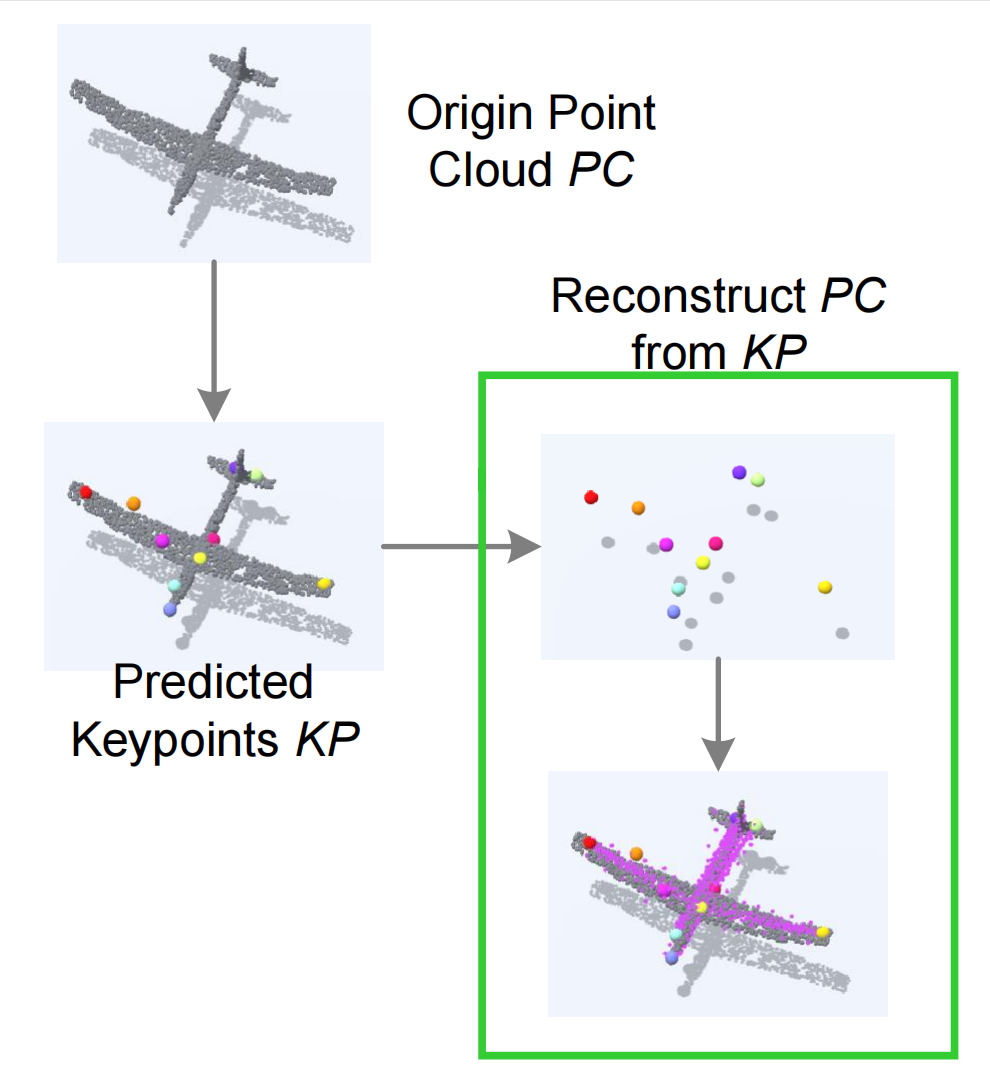}
    }\subfloat[Our Method]{\includegraphics[width =.53\linewidth]{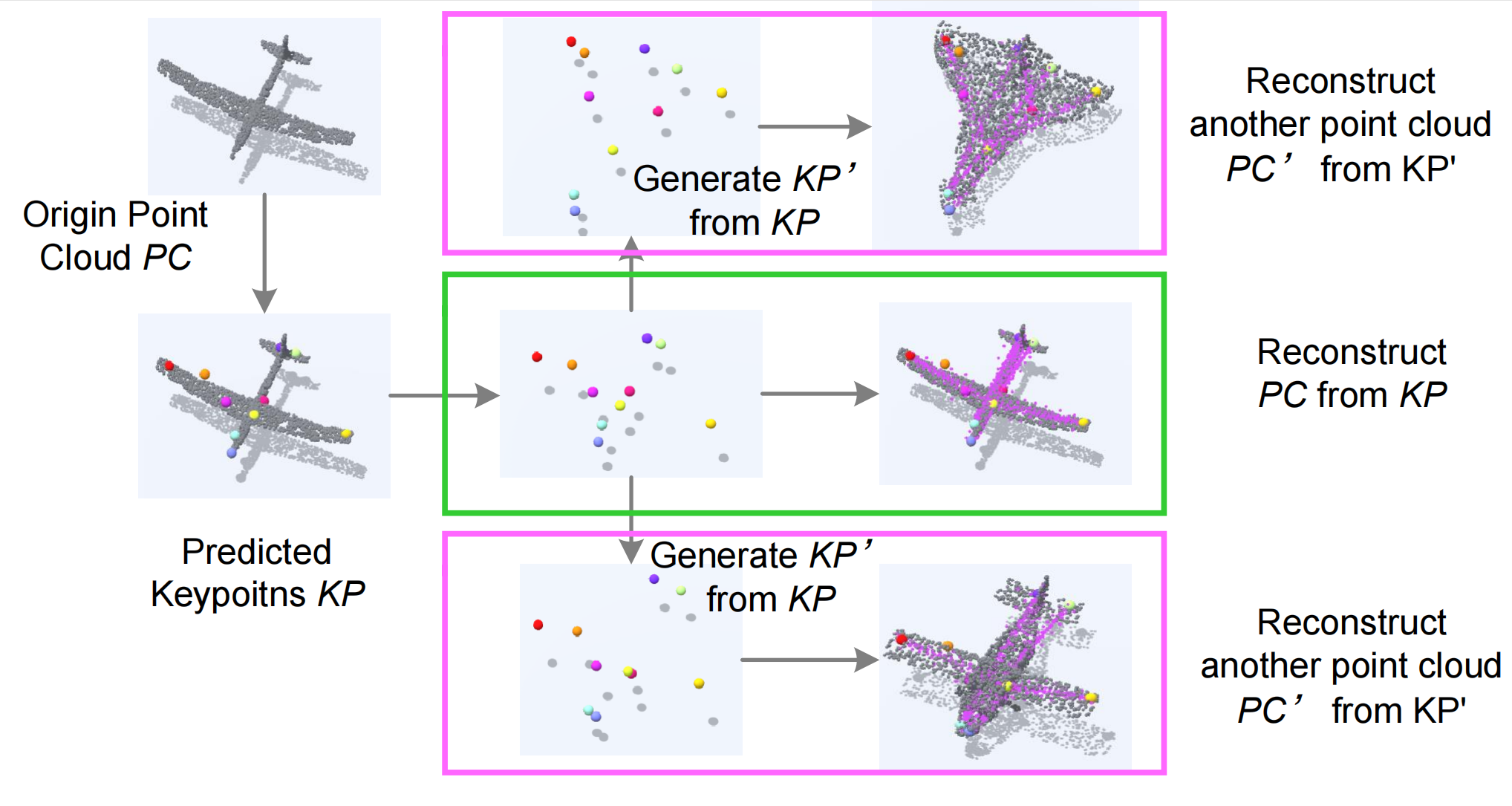}}
    \caption{{\textbf{Comparison of our method and previous methods}}. Previous methods focus on self-reconstruction, which may fail to mine category-level semantic consistency information. We address this issue with mutual reconstruction (the keypoints of an object also reconstruct other objects in the same category).}
    \label{figure1 novelty}
\end{figure}

\end{abstract}

\section{Introduction}
3D semantic keypoints generally refer to representative points on 3D objects, which possess category-level semantic consistency through categories.
Detecting 3D semantic keypoints has a broad application scenarios, such as 3D registration~\cite{shape_matching1}, 3D reconstruction~\cite{c3dpo}, shape abstraction~\cite{sa1} and deformation~\cite{kpd}. However, this task is quite challenging because of unknown shape variation among different instances in a category, unordered point cloud representations, and limited data annotations.

From the technical view, 3D keypoint detection can be divided into geometry-only~\cite{harris3D,ISS} and learning-based~\cite{kpd,usip,skeleton,symmetric,instrinsic}. For geometry-only ones, they generally leverage shape attributes such as normals to detect distinctive and repeatable points, however, they generally fail to mine semantic information. Learning-based methods can learn semantics from massive training data and can be further classified into supervised and unsupervised. As illustrated in previous works~\cite{kpd,skeleton}, supervised methods may suffer from limited human annotated data~\cite{keypointnet}, which greatly limits their applicability. Unsupervised learning of 3D semantic keypoints~\cite{skeleton,usip,kpd}, however, is particularly challenging due to the ambiguity of semantic information when labels are not given. A few trails have been made toward this line, and we divide these unsupervised methods into two classes by examining if the method employs category-level information explicitly or implicitly. 1) Implicit methods focus on self-related tasks of a single object, such as self-reconstruction~\cite{skeleton,instrinsic}, where keypoints of each object are optimized to reconstruct the original object; category-level information is ensured in an indirect way, as all objects in a specific category are fed into the model during the training process. 2) There are only a few explicit methods~\cite{kpd,symmetric}, which consider category information directly. The networks are usually driven by losses of specific tasks involving more than one object from a category. Both explicit and implicit methods have made great success in terms of geometric consistency and robustness, but still fail to ensure semantic consistency. For the implicit methods~\cite{skeleton,instrinsic}, this is caused by a lack of semantic information, as they only consider a single object in a whole category, e.g., reconstructing the object itself based on its own keypoints. As for explicit methods~\cite{kpd,symmetric}, although category-level information are taken into consideration explicitly, they still tend to pursue consistency and fail to mine the hidden semantic information within keypoints.

To this end, from a novel mutual reconstruction perspective, we propose an unsupervised method to learn category-level semantic keypoints from point clouds. {\textit{We believe that semantic consistent keypoints of an object should be able to reconstruct itself as well as other objects from the same category.}} The motivation behind is to fully leverage category-level semantic information and ensure the consistency based on an explicit manner. Compared with deformation tasks~\cite{kpd} based on cage deformation methods, shape reconstruction from keypoints have been well investigated~\cite{skeleton,instrinsic} and is more straightforward and simpler. In particular, only reconstruction task is involved in our model. The overall technique pipeline of our method is as follows. First, given two point clouds of the same category, keypoints are extracted by an encoder; second, the source keypoint set is reshaped according to the offset of input point clouds; then, source and reshaped keypoint sets are used as the guidance for self-reconstruction and mutual reconstruction with a decoder~\cite{skeleton}; finally, both self-reconstruction and mutual reconstruction losses are considered to train the network. Experimental results on KeypointNet~\cite{keypointnet} and ShapeNet Part~\cite{shapenet} have shown that the proposed model outperforms the state-of-the-arts on human annotation datasets. It can be also generalized to real-world scanned data~\cite{dai2017scannet} without human annotations. 

Overall, our method has two key contributions:  
\begin{itemize}
    \item To the best of our knowledge, we are the first to mine semantic consistency with mutual reconstruction, which is a simple yet effective way to detect consistent 3D semantic keypoints.
    \item We propose a network to ensure keypoints performing both self reconstruction and mutual reconstruction. It achieves the overall best performance under several evaluation metrics on KeypointNet~\cite{keypointnet} and ShapeNet Part~\cite{shapenet} datasets.
\end{itemize}

\section{Related Work}
This section first gives a review on unsupervised semantic keypoints and geometric keypoint detection. Supervised methods are not included, since the task of 3D keypoint detection is seldomly accomplished in a supervised way due to the lack of sufficient labelled datasets. Then, a recap on deep learning on point clouds is given.
\subsection{Unsupervised Semantic Keypoint Detection}
We divide current methods into two classes according to if the category-level information is leveraged implicitly or explicitly.
\subsubsection{Implicit methods.} Implicit methods employ self-related metrics to measure the quality of keypoints. A typical implicit method is skeleton merger~\cite{skeleton}, whose key idea is to reconstruct skeleton-liked objects based on its keypoints through an encoder-decoder architecture. Another implicit way~\cite{instrinsic} utilizes a convex combination of local points to generate local semantic keypoints, which are then measured by how close they are to the origin point cloud. Unsupervised stable interest point (USIP)~\cite{usip} predicts keypoints with a Siamese architecture, and the two inputs are two partial views from a 3D object. Implicit methods can achieve good spatial consistency and are relatively light-weight. However, they generally fail to mine semantic consistency information. 
\subsubsection{Explicit methods.} 
% We cast this as the problem of aligning a source 3D object to a target 3D object from the same object category. Our method analyzes the difference between the shapes of the two objects by comparing their latent representations. This latent representation is in the form of 3D keypoints that are learned in an unsupervised way. The difference between the 3D keypoints of the source and the target objects then informs the shape deformation algorithm that deforms the source object into the target object.

Explicit methods cope with category-level information directly. Keypoint deformer~\cite{kpd} employs a Siamese architecture for shape deformation to detect shape control keypoints; the difference between two input shapes is analysed by comparing their keypoint sets. The cage~\cite{cage} method is crucial to keypoint deformer~\cite{kpd}; to deform a point cloud, cage~\cite{cage} takes the origin point cloud, shape control points on point cloud, and target cage as input, the output of cage consists of a deformed cage and a deformed point cloud under the constraint of cage. Another explicit method~\cite{symmetric} learns both category-specific shape basis and instance-specific parameters such as rotations and coefficients during training; however, the method requires a symmetric hypothesis. Different from the two previous works, our method evaluates keypoints from the self and mutual reconstruction quality by estimated keypoints and do not require additional hypotheses on inputs.
\subsection{Geometric Keypoint Detection}
Besides semantic keypoints, detection of geometric keypoints has been well investigated in previous works~\cite{harris3D,ISS}. Different from semantic keypoints that focus on category-level semantic consistency, geometric keypoints are defined to be repeatable and distinctive keypoints on 3D surfaces. In a survey on geometric keypoints, Tombari et al.~\cite{tombari2013performance} divided 3D geometric detectors into two categories, i.e., fixed-scale and adaptive-scale. Fixed-scale detectors, such as LSP~\cite{lsp}, ISS~\cite{ISS}, KPQ~\cite{KPQ} and HKS~\cite{HKS}, find distinctive keypoints at a specific scale with a non-maxima suppression (NMS) procedure, which is measured by saliency. Differently, adaptive-scale detectors such as LBSS~\cite{LBSS} and MeshDoG~\cite{MeshDoG}  first build a scale-space defined on the surface, and then pick up distinctive keypoints with an NMS of the saliency at the characteristic scale of each point. Geometric keypoints focus on repeatable and distinctive keypoints rather than semantically consistent keypoints. 
\subsection{Deep Learning on Point Clouds}
Because our method relies on reconstruction, which is typically performed with an encoder-decoder network on point clouds. We will briefly discuss deep learning methods from the perspectives of encoder and decoder.
\subsubsection{Encoder.} A number of neural networks have been proposed, e.g., PointNet~\cite{pointnet}, PointNet++~\cite{pointnet++}, and PointConv~\cite{pointconv}, which directly consume 3D point clouds. PointNet~\cite{pointnet} is a pioneering work, which extracts features from point clouds with point-wise MLPs and permutation-invariant functions. Based on PointNet, PointNet++~\cite{pointnet++} introduces a hierarchical structure to consider both local and global features; PointNet is applied after several sampling and grouping layers; PointNet++ is also employed as an encoder by several unsupervised 3D semantic keypoint detection methods~\cite{skeleton,instrinsic}. More recent point cloud encoders include~\cite{simonovsky2017dynamic,pointconv,wang2019dynamic}. These encoders have achieved success in tasks like registration~\cite{shape_matching1} and reconstruction~\cite{c3dpo}. Several keypoint detection methods\cite{kpd,skeleton,instrinsic} also employ PointNet++~\cite{pointnet++} as the encoder.

\subsubsection{Decoder.}
In previous point cloud learning works~\cite{achlioptas2018learning,yu2018pu}, MLP is frequently leveraged to generate point clouds from the encoded features. Specifically, FoldingNet~\cite{foldingnet} proposes a folding-based decoder to deform 2D grid onto 3D object surface of a point cloud. Many works~\cite{deng2018ppf,deprelle2019learning,yang2019pointflow} follow FoldingNet~\cite{foldingnet} and decode the features based on structure deformation. In~\cite{shu20193d}, tree structure is used to decode structured point clouds. 
From the functional view, most decoders leveraged by 3D semantic keypoint detection methods~\cite{skeleton,usip,instrinsic,symmetric} focus on reconstructing the original shape of the input. An exception is keypoint deformer~\cite{kpd}, whose decoder tries to deform the source shape into the target shape through cage-based deformation.
\section{The Proposed Method}
\begin{figure}[ht]
    \centering
    \includegraphics[width=.8\linewidth]{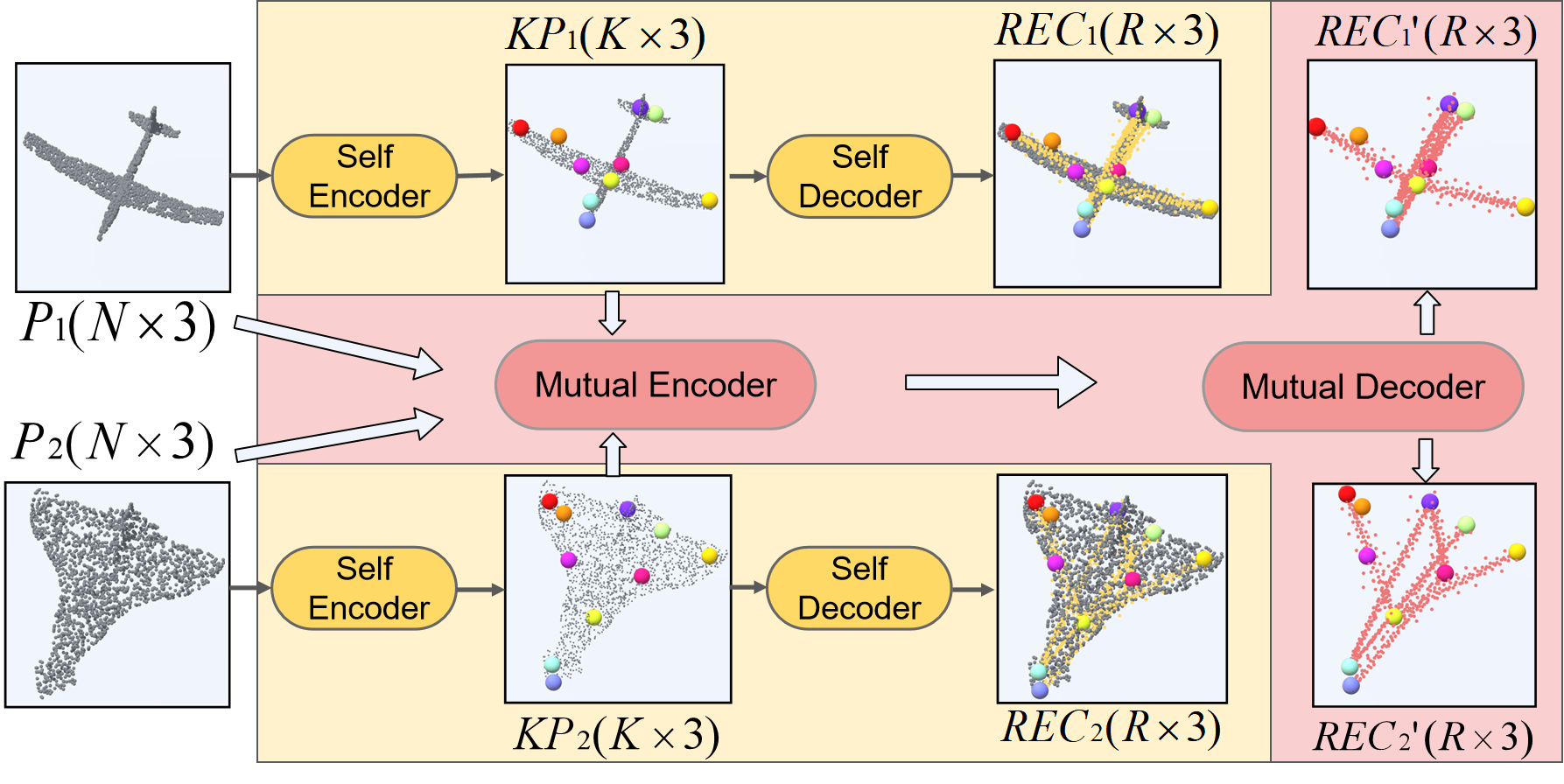}
    \caption{\textbf{Pipeline of our method.} Two input point clouds (each with $N$ points) $P_1, P_2$ are fed into self and mutual encoders, the outputs are two keypoint sets $KP_1$, $KP_2$ and mutual features. Self and mutual decoders then decode the source keypoint set $KP_1, KP_2$  into  $REC_1, REC_2$ and $REC_1', REC_2'$. Reconstruction loss is calculated by Chamfer distance between $P, REC$ (self reconstruction) and $REC, REC'$ (mutual reconstruction).}
    \label{pipeline}
\end{figure}

The pipeline of our method is shown in Fig.~\ref{pipeline}. Self reconstruction and mutual reconstruction are performed simultaneously through encoder-decoder architectures.
\subsection{Self and Mutual Reconstructions}
Self and mutual reconstructions are the key components of our method. For an input point cloud $P_1$, self-reconstruction is supposed to reconstruct the origin point cloud $P_1$ from its own keypoint set $KP_1$; mutual reconstruction is to reconstruct another point cloud $P_2$  with $KP_1$ and the offset between $P_1, P_2$. 

\subsubsection{Mutual reconstruction.}
\begin{figure}[t]
    \centering
    \includegraphics[width=.8\linewidth]{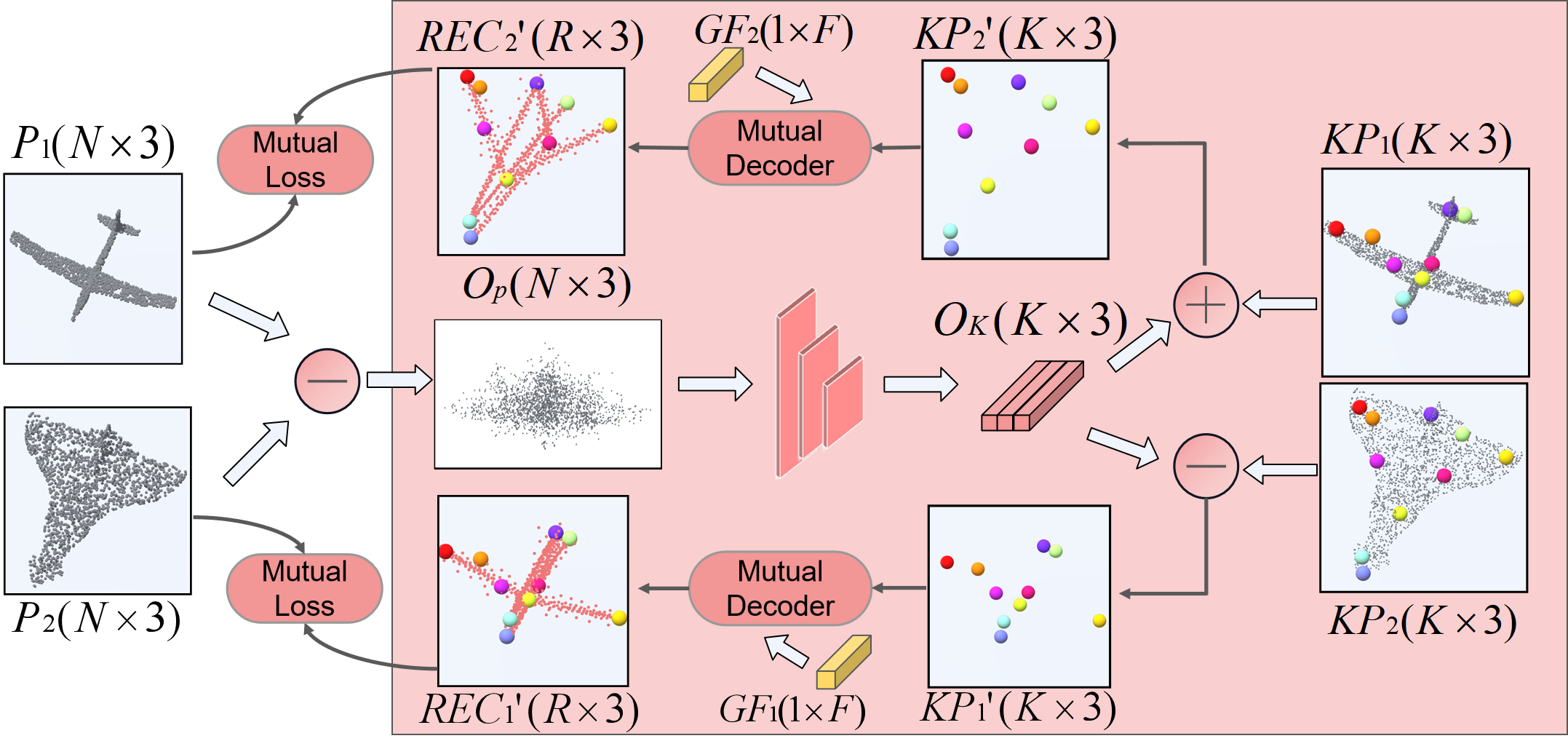}
    \caption{\textbf{Mutual reconstruction process of our method.} Two predicted keypoint sets $KP_1$ and $KP_2$ are reshaped into $KP_1'$ and $KP_2'$ with offsets generated between input point clouds $PC_1, PC_2$. A \textbf{shared} decoder from Skeleton Merger~\cite{skeleton}, then decodes $KP_1', KP_2'$ into $REC_1', REC_2'$. Mutual reconstruction loss is calculated by Chamfer distance between $REC, REC'$.}
    \label{fig:mutual recon}
\end{figure}

Our mutual reconstruction module is depicted in Fig.~\ref{fig:mutual recon}. The mutual reconstruction process utilizes several outputs from self reconstruction, including keypoint sets $KP_1, KP_2$ and the global feature $GF$. 

\begin{figure}[ht]
    \centering
    \subfloat[Self reconstruction]{\includegraphics[width=.75\linewidth]{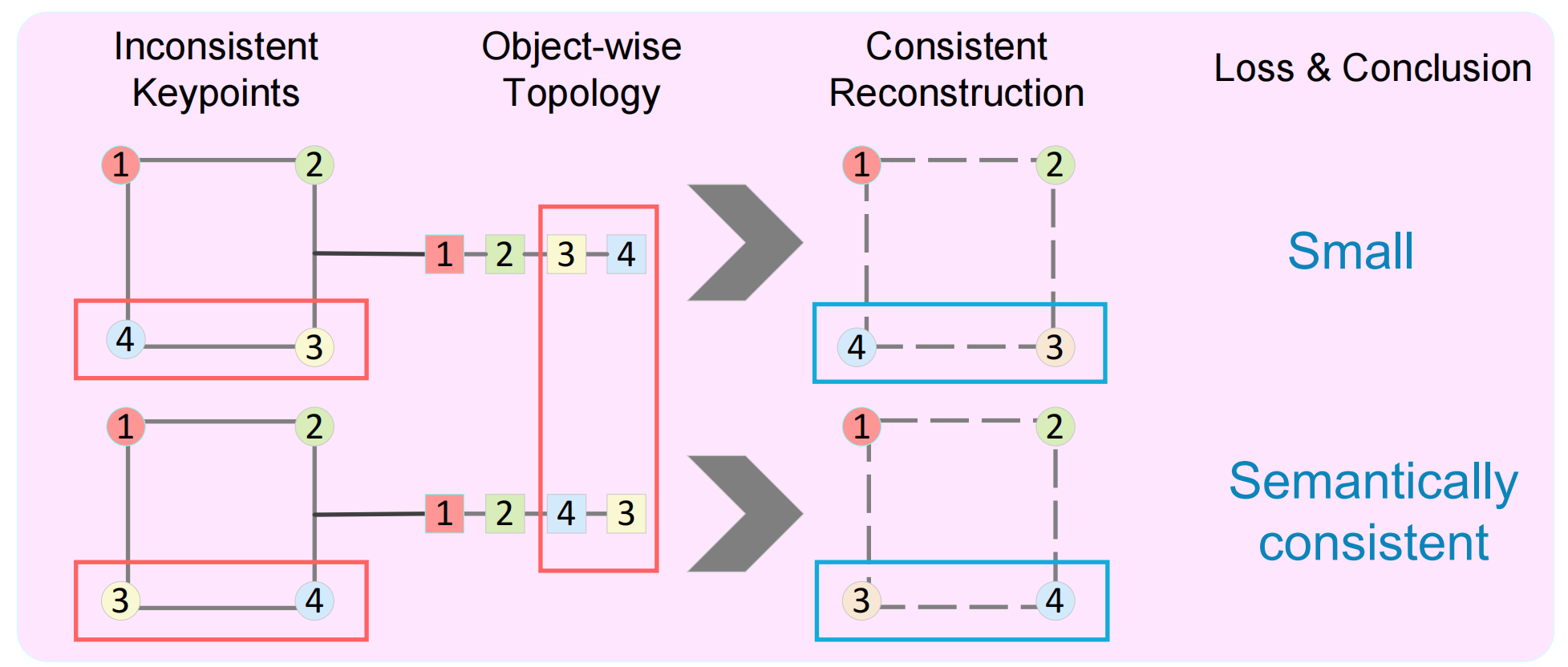}}
    
    \subfloat[Mutual reconstruction]{\includegraphics[width=.75\linewidth]{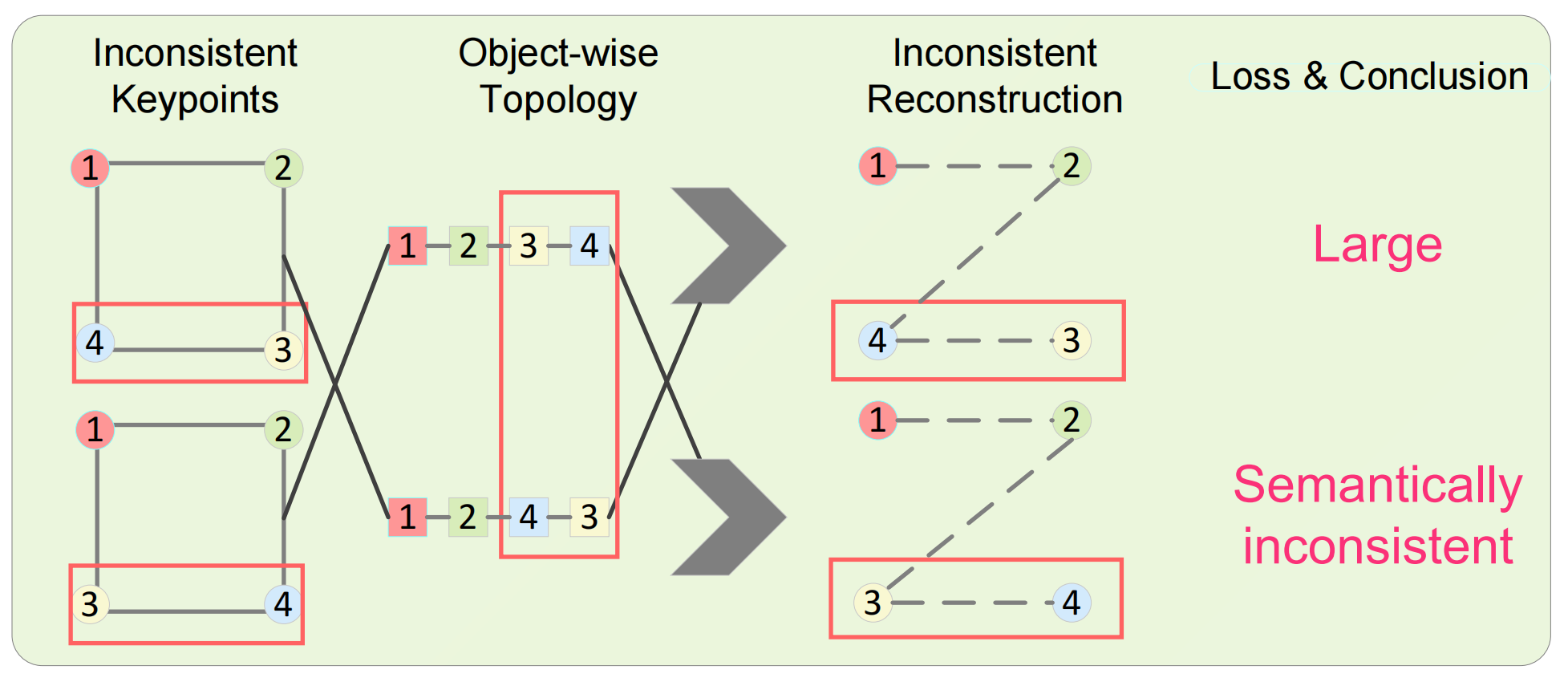}}
    \caption{\textbf{Illustration of the difference between self-related tasks and mutual reconstruction.} The predictor outputs inconsistent semantic labels 3 \& 4 for two objects from the same category (square) in self-reconstruction; the misalignment may be ignored as the model can learn to output different topology information (e.g., 1-2-4-3 for misaligned 3 \& 4). By contrast, in mutual reconstruction, such inconsistency issue can be identified, resulting in an obvious large reconstruction loss. }
    \label{fig:self_and_mutual_compare}
\end{figure}

Mutual reconstruction is supposed to be able to extract category-level semantic consistent keypoints as illustrated in Fig.~\ref{fig:self_and_mutual_compare}. The figure illustrates the semantic ambiguity of self-reconstruction, which can be resolved by the mutual reconstruction module. When the method with only self-related tasks (e.g., self reconstruction) predicts object-wise keypoints which are not semantically consistent, it may fail to notice the inconsistency as the topology information is inconsistent as well (we visualize the topology information as a sequence of connection, while some methods employ topology information implicitly); however, the mutual reconstruction model is sensitive when either the topology or semantic label prediction is not correct, as additional shapes are considered in mutual reconstruction and the constraint on keypoint consistency is much tighter.

% additional shapes are utilized to examine each other.

\subsubsection{Self reconstruction.}
\begin{figure}[t]
    \centering
    \includegraphics[width=.8\linewidth]{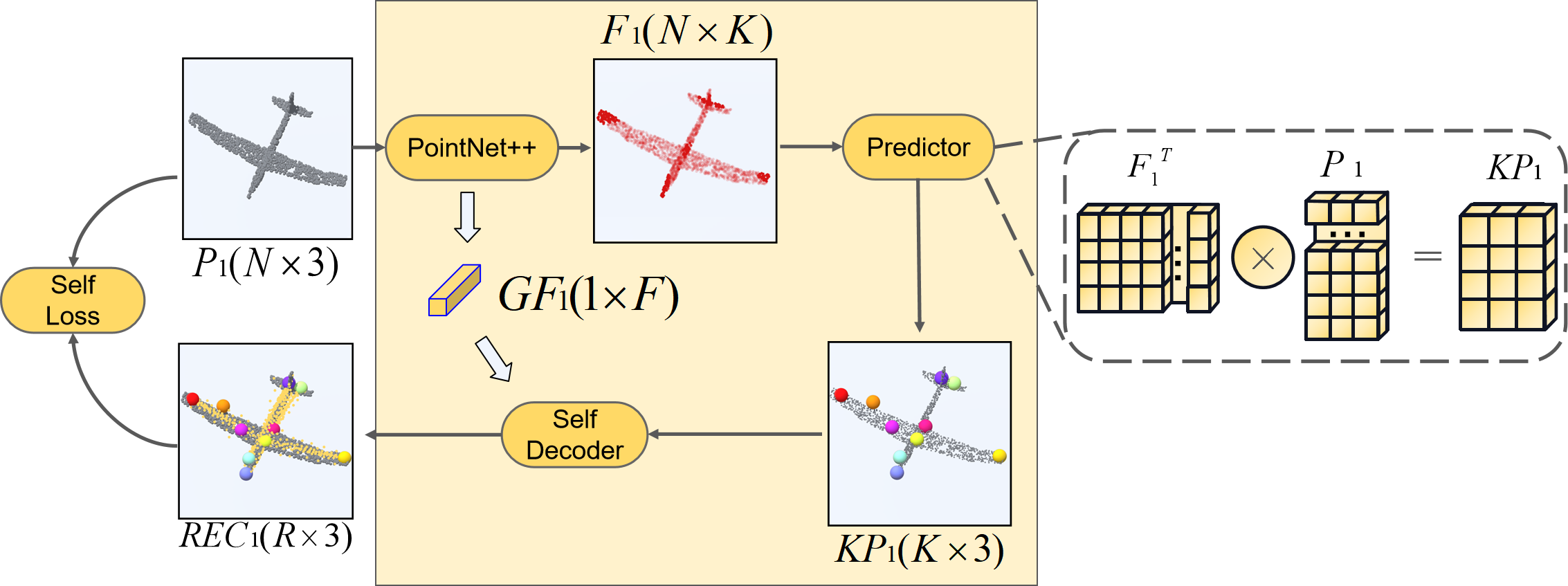}
    \caption{\textbf{Self reconstruction process of our method.} The input point cloud (with $N$ 3D points) $P_1$ is first fed into a shared PointNet++ encoder, whose output is a group of $K\times N$ point-wise feature $F_1$, where $K$ indicates the number of keypoints. Keypoint sets $KP_1$ is calculated by the inner product of $F_1^T$ and $P$. A shared decoder then reconstructs the source keypoint $KP_1$ sets into $REC_1$. Self reconstruction loss is calculated by Chamfer distance between $P_1, REC_1$. The $GF_1$ indicates the global feature, which is consists of activation strengths and trainable offsets.}
    \label{fig:self recon}
\end{figure}

The self reconstruction module is presented in Fig.~\ref{fig:self recon}. Specifically, the point-wise feature can also be considered as point-wise score, because the keypoints are actually generated by linear combination of origin points. In other words, for the point with a higher score (feature value), it contributes more to the keypoint prediction. We simply define the point-wise score to be the sum of the $k$-dim feature, as visualized in Fig.~\ref{fig:mutual recon} (the airplane in red).

Self reconstruction is also a critical component for mining the semantic information from an instance~\cite{skeleton,instrinsic}. To ensure category-level semantic consistent information, instance and cross-instance information should be mined, such that self reconstruction is utilized as complementary to mutual reconstruction.
\subsection{Network Architecture}
The whole pipeline of our method is illustrated in Fig.~\ref{pipeline}. All \textbf{decoders} in self and mutual reconstruction processes are shared, and the only difference between the self and mutual \textbf{encoder} is that the mutual one needs to \textbf{reshape keypoint set} after the same architecture as the self one.  Thus, the core of our network architecture are encoder, reshaping keypoint set and decoder. The three technical modules are detailed in the following.
\subsubsection{Encoder.}
The designed encoder is supposed to generate keypoints proposals $K_1, K_2$ from input point clouds $P_1, P_2$. First, we employ the PointNet++~\cite{pointnet++} encoder and it offers a $K$-dimension point-wise feature for every point in the origin point cloud, thus the shape of point feature matrix $F$ is $K\times N$. Keypoints are calculated by:
\begin{equation}
    KP = F \cdot P.
\end{equation}

\subsubsection{Reshape keypoint set.}
After keypoints proposals $KP$ are generated by the encoder, they are reshaped into new keypoint sets $KP'$, which are utilized by the decoder for mutual reconstruction. We reshape source keypoint set $KP$ with a point-wise offsets $O_{kp}$ as:
\begin{equation}
    KP'_1 = KP_2 + O_{K},
\end{equation}
and
\begin{equation}
    KP'_2 = KP_1 - O_{K},
\end{equation}
where $O_{kp}$ is calculated by feeding offsets of origin point clouds $O_p$ into a 3-layers MLP, as in the following:
\begin{equation}
    O_{K} = MLP(O_P),
\end{equation}
and
\begin{equation}
    O_P = K_1 - K_2.
\end{equation}
The reshaped source keypoints are fed to the decoder for reconstruction.
\subsubsection{Decoder.}
We build our decoder following skeleton merger~\cite{skeleton}. The decoder takes keypoint sets $KP, KP'$ and global feature (activation strengths and trainable offsets) as input. It first generates $n(n-1)/2$ line-like skeletons, each of them is composed of a series of points with fixed intervals. Second, trainable offsets are added to every point on the skeleton-like point cloud. Finally, $n(n-1)/2$ activation strengths are applied to the $n(n-1)/2$ skeletons for reconstruction; only skeletons with high activation strengths contribute to the reconstruction process. As such, shapes are reconstructed by decoders.

\subsection{Losses and Regularizers}
Both self and mutual reconstruction losses are employed to train our model in an unsupervised way.
\subsubsection{Reconstruction losses.}
We calculate reconstruction loss with Composite Chamfer Distance (CCD)~\cite{skeleton}. CCD is a modified Chamfer Distance which takes the activation strengths into consideration. For fidelity loss, the CCD between $\hat{X}$ and $X$ is given as:
\begin{equation}
    L_{f}=
    \sum_i a_i \sum_{\hat{p}\in \hat{X}_i}\min_{p_0\in X}\lVert {\hat{p}-p_0}\rVert_2,
\end{equation}
where $\hat{X}_i$ is the $i$-th skeleton of point cloud $\hat{X}$, and $a_i$ is the activation strength of $\hat{X}_i$. For the coverage loss, there is a change from the fidelity loss that more than one skeleton are considered in the order of how close they are to the given point, until the sum of their activation strengths exceeds $1$~\cite{skeleton}.

We apply the CCD loss in both self and mutual reconstruction tasks. For self reconstruction, we calculate the CCD between the input target shape $P_t$ and output target shape $P_t'$:
\begin{equation}
    L_{rec_{s}}=
    CCD(P_1, REC_1) + CCD(P_2, REC_2).
\end{equation}
For mutual reconstruction, we calculate the CCD between the input target shape $P_t$ and output source shape $P_s'$:
\begin{equation}
    L_{rec_{m}}=
    CCD(P_1, REC_1') + CCD(P_2, REC_2').
\end{equation}

The eventual reconstruction loss is a combination of the two losses:
\begin{equation}
    L_{rec} = \lambda_{s}L_{rec_{s}}+\lambda_{m}L_{rec_{m}},
\end{equation}
where $\lambda_{s}$ and $\lambda_{m}$ are weights to control the contributions of self and mutual reconstructions.

\subsubsection{Regularizers.}
The trainable offsets in our decoder are calculated by multiple MLPs. To keep the locality of every points on the skeleton, we apply an $L_2$ regularization on them. $L_2$ regularization is also imposed on the keypoint offset $O_{K}$, in order to reduce the geometric changes of keypoints when reconstructing the other shape.

\section{Experiments}
\subsubsection{Experimental setup.} In our experiments, we follow~\cite{skeleton} and report the dual alignment score (DAS),  mean intersection over union (mIoU), part correspondence ratio, and robustness scores of tested methods. We choose learning-based methods including skeleton merger~\cite{skeleton}, Fernandez et al.~\cite{symmetric}, USIP~\cite{usip}, and D3Feat~\cite{bai2020d3feat}; and geometric methods including ISS~\cite{ISS}, Harris3D~\cite{harris3D}, and SIFT3D~\cite{sift}, for a thorough comparison. Note that there is a lack of supervised methods~\cite{yumer2015semantic} and valid annotated datasets~\cite{keypointnet} in this field. For this reason, only several unsupervised ones are chosen. We also perform an ablation study, in which we analyze the effectiveness of our mutual-reconstruction module. For training, We employ ShapeNet~\cite{shapenet} with the standard split of training and testing data, in which all shapes are normalized into a unit box. For evaluation, we utilize the following datasets, i.e., the human-annotated keypoint dataset KeypointNet~\cite{keypointnet}, a part segmentation dataset named ShapeNet Part~\cite{shapenet}, and a real-world scanned dataset ScanNet~\cite{dai2017scannet}.
%We conduct several experiments to evaluate our method. Generally, we hope our method can predict semantic keypoints that are semantically consistent and close to human intuition, so we follow~\cite{skeleton} and calculate the DAS score, mIoU, part correspondence ratio and robustness score of our method. We also need to figure out the effectiveness of our mutual-reconstruction module, which is the core of our method. To achieve this, ablation study is performed where we compare the self-and-mutual reconstruction structure with self-only reconstruction structure. 

%To make it comprehensive, we choose both learning-based methods(Skeleton Merger~\cite{skeleton}, the method of Fernandez et al.~\cite{symmetric}, USIP~\cite{usip}, D3Feat~\cite{bai2020d3feat}) and tradition 3D keypoint detectors(ISS~\cite{ISS}, Harris3D~\cite{harris3D}, SIFT3D~\cite{sift}) for comparison. There is a lack of supervised methods~\cite{yumer2015semantic} and valid annotated datasets~\cite{keypointnet} in this field. For that reason, only several unsupervised ones are chosen.

\subsubsection{Implementation details.} We randomly split the training dataset into two groups. Mutual reconstruction is performed by respectively taking two shapes from two different groups per time. Point clouds are down-sampled to 2048 points with farthest-point-sampling\cite{pointnet++}. The number of keypoints for all categories is restricted to 10. The model is trained on a single NVIDIA GTX 2080Ti GPU, and the Adam~\cite{adam} is used as the optimizer. We train the KeypointNet\cite{keypointnet} for 80 epochs in 8 hours. By default, the weights ($\lambda_{s}$ and $\lambda_{m}$) of self reconstruction and mutual reconstruction losses are set to 0.5.
%\subsubsection{Dataset.}

\subsection{Semantic Consistency}
Semantic consistency means a method can predict keypoints that are of the same semantic information. There are several popular metrics to evaluate semantic consistency, all of which are considered in the experiments for a comprehensive evaluation.
\subsubsection{Dual alignment score.}
We first evaluate the semantic consistency on KeypointNet\cite{keypointnet} with DAS, which is introduced by~\cite{skeleton}. Given the estimated keypoints on a source point cloud, DAS employs a reference point cloud for keypoint quality evaluation. We predict keypoints with our model on both source and reference point clouds, and use the human annotation on the reference point cloud to align our keypoints with annotated keypoints. The closet keypoint to a human annotation point is considered to be aligned with the annotation. DAS then calculates the ratio of of aligned keypoints between the source and reference point clouds.

The results are shown in Table~\ref{tab:DAS}. It can be found that ISS is significantly inferior to others, because it tries to find distinctive and repeatable points rather than points with semantic information. Compared with two recent unsupervised learning methods, our method also surpasses them in most categories.

\begin{table}[t]
    \centering
    \caption{Comparative DAS performance on KeypointNet.}
    \begin{tabular}{c|c cc cc cc c|c}
\toprule
    & Airplane& Chair& Car & Table & Bathtub&Guitar&Mug&Cap&Mean\\
\midrule
Fernandez et al.~\cite{symmetric}& 61.4& 64.3&--&--&--&--&--&--&62.85\\
Skeleton Merger~\cite{skeleton}& 77.7& 76.8&\textbf{79.4}&70.0&69.2&\textbf{63.1}&67.2&53.0&69.55\\
ISS~\cite{ISS}&13.1&10.7&8.0&16.2&9.2&8.7&11.2&13.1&11.28\\
\midrule
Ours& \textbf{81.0}& \textbf{83.1}&74.0&\textbf{78.5}&\textbf{71.2}&61.3&\textbf{68.2}&\textbf{57.1}&\textbf{71.8}\\
\bottomrule
    \end{tabular}
    
    \label{tab:DAS}
\end{table}
\subsubsection{Mean intersection over union.}
We report the mIoU of predicted keypoints and ground truth ones. The results are shown in Table~\ref{tab:mIoU}. 

As witnessed by the table, our method achieves the best performance on four categories. Noth that Fernandez et al.~\cite{symmetric} only reported results on the `Airplane' and `Chair' categories, while our method still outperforms it on these two categories.
\begin{table}[t]
    \centering
    \caption{Comparative mIoU performance on KeypointNet.}
    \begin{tabular}{c|cccccc|c}
\toprule
    & Airplane & Chair & Car &Table &Bed &Skateboard &Mean\\
\midrule
Fernandez et al.~\cite{symmetric}& 69.7 & 51.2 &--&--&--&--&--\\
Skeleton Merger~\cite{skeleton}&\textbf{79.4}&68.4&47.8&50.0&\textbf{47.2}&40.1&55.48\\
ISS~\cite{ISS}&36.3&11.6&20.3&24.1&33.7&31.0&26.16\\

\midrule
Ours &79.1 &\textbf{68.9}&\textbf{51.7}&\textbf{54.1}&45.4&\textbf{43.3}&\textbf{57.08}\\

\bottomrule
    \end{tabular}
    
    \label{tab:mIoU}
\end{table}

\subsubsection{Part correspondence ratio.}
We also test the mean part correspondence ratio on ShapeNet Part dataset. This metric is not as strict as DAS and mIoU, because it defines two semantic keypoints are corresponding if they are in the same semantic part of objects in a category. The comparative results are shown Table~\ref{tab:correspo}.

Due the that the part correspondence ratio is a loose metric, the gaps among tested methods are not as dramatic as those in Tables~\ref{tab:DAS} and~\ref{tab:mIoU}. Remarkably, our method also achieves the best performance under this metric.
\begin{table}[t]
    \caption{Mean correspondence ratio results on ShapeNet part dataset.}
    \centering
    \begin{tabular}{c|ccc|c}
    \toprule
         &Airplane &Chair & Table &Mean \\
    \midrule
    USIP~\cite{usip} &77.0 & 70.2 & 81.5 &76.23\\
    D3Feat~\cite{bai2020d3feat} & 79.9 &84.0 &79.1 &81.00\\
    \midrule
    Harris3D~\cite{harris3D} &76.9 &70.3& 84.2&77.13\\
    ISS~\cite{ISS}& 72.2&68.1&83.3&74.53\\
    SIFT3D~\cite{sift}&73.5&70.9&84.1&76.17\\
    \midrule
    Ours&\textbf{81.5}&\textbf{85.2}&\textbf{85.7}&\textbf{84.13}\\
    \bottomrule
    \end{tabular}
    \label{tab:correspo}
\end{table}
\subsubsection{Visualization.} 
\begin{figure}[h]
    \centering
    \subfloat[3D coordinates distribution]{
    \includegraphics[width=.48\linewidth]{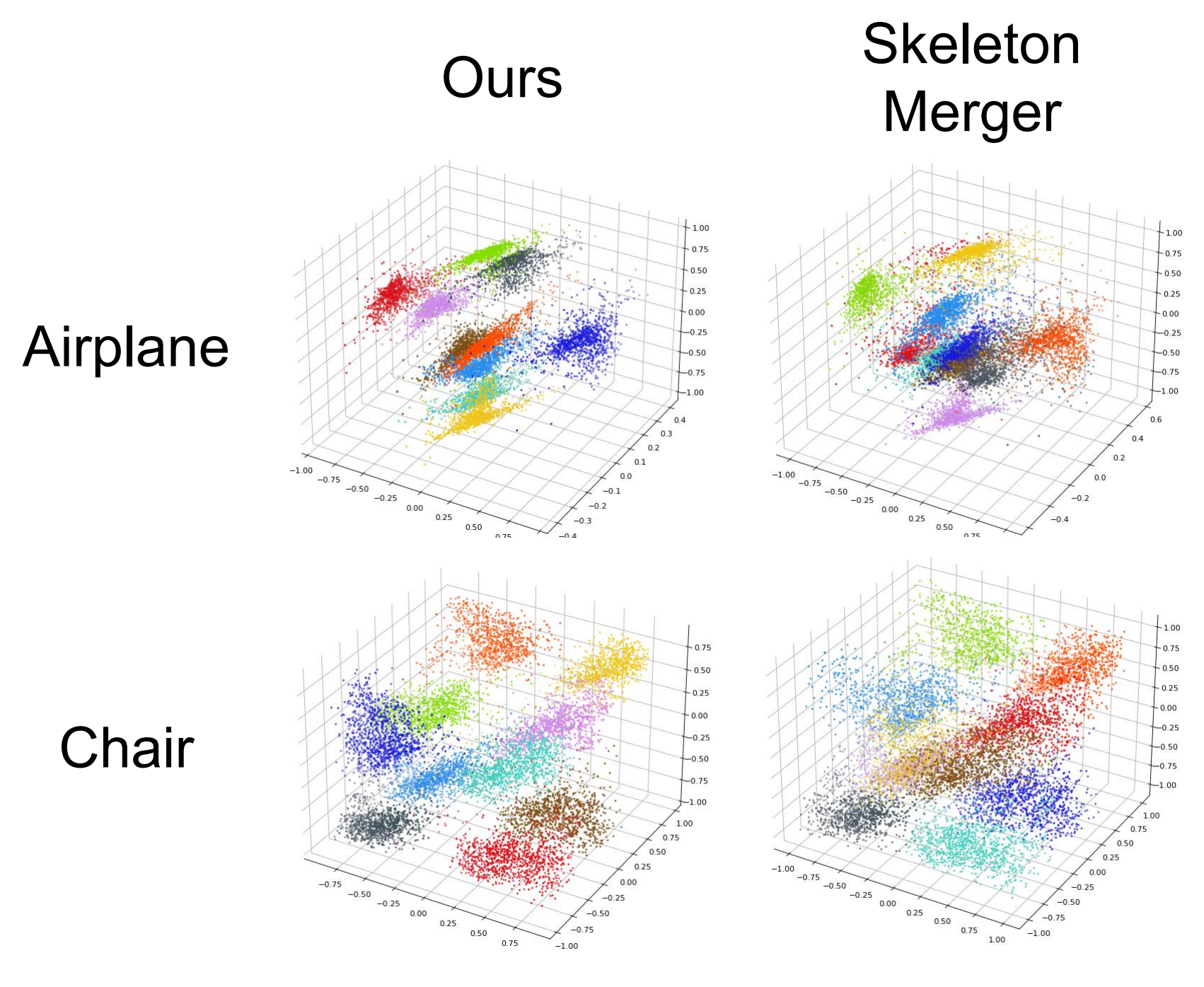}
    }\subfloat[2D t-SNE feature distribution]{
    \includegraphics[width=.34\linewidth]{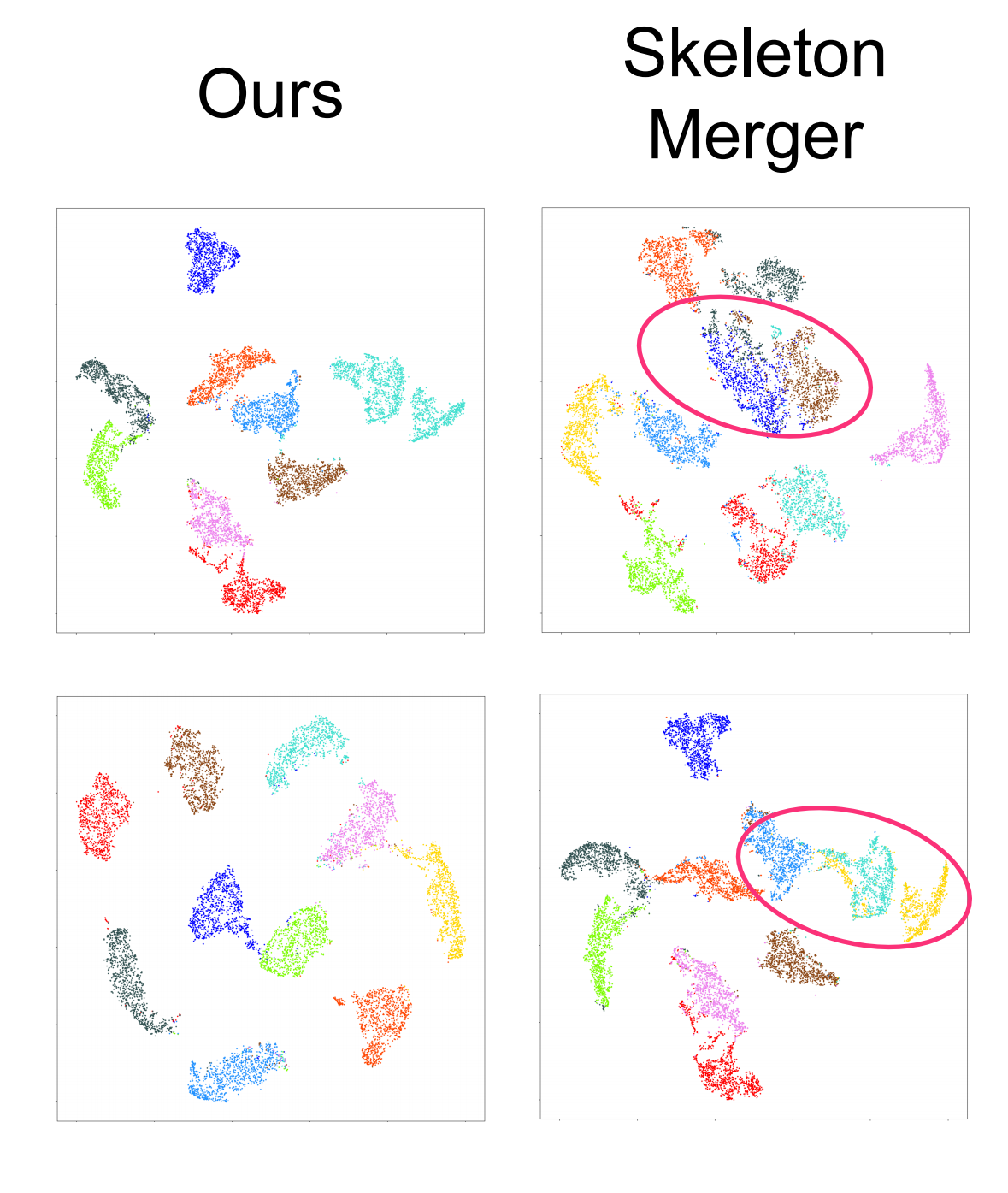}
    }
    \caption{\textbf{Distribution of semantic keypoints in the 3D space and keypoint features in the 2D space with t-SNE.} Points with the same semantic label are rendered with the same color.}
    \label{fig:visualization of featrues}
\end{figure}

We first visualize the 3D keypoint distribution and the keypoint features based on t-SNE in Fig.~\ref{fig:visualization of featrues}, where different colors indicate different semantic labels. Here, we take the skeleton merger method as a comparison.

\begin{figure}[t]
    \centering
    \includegraphics[width=.68\linewidth, height=5.7cm]{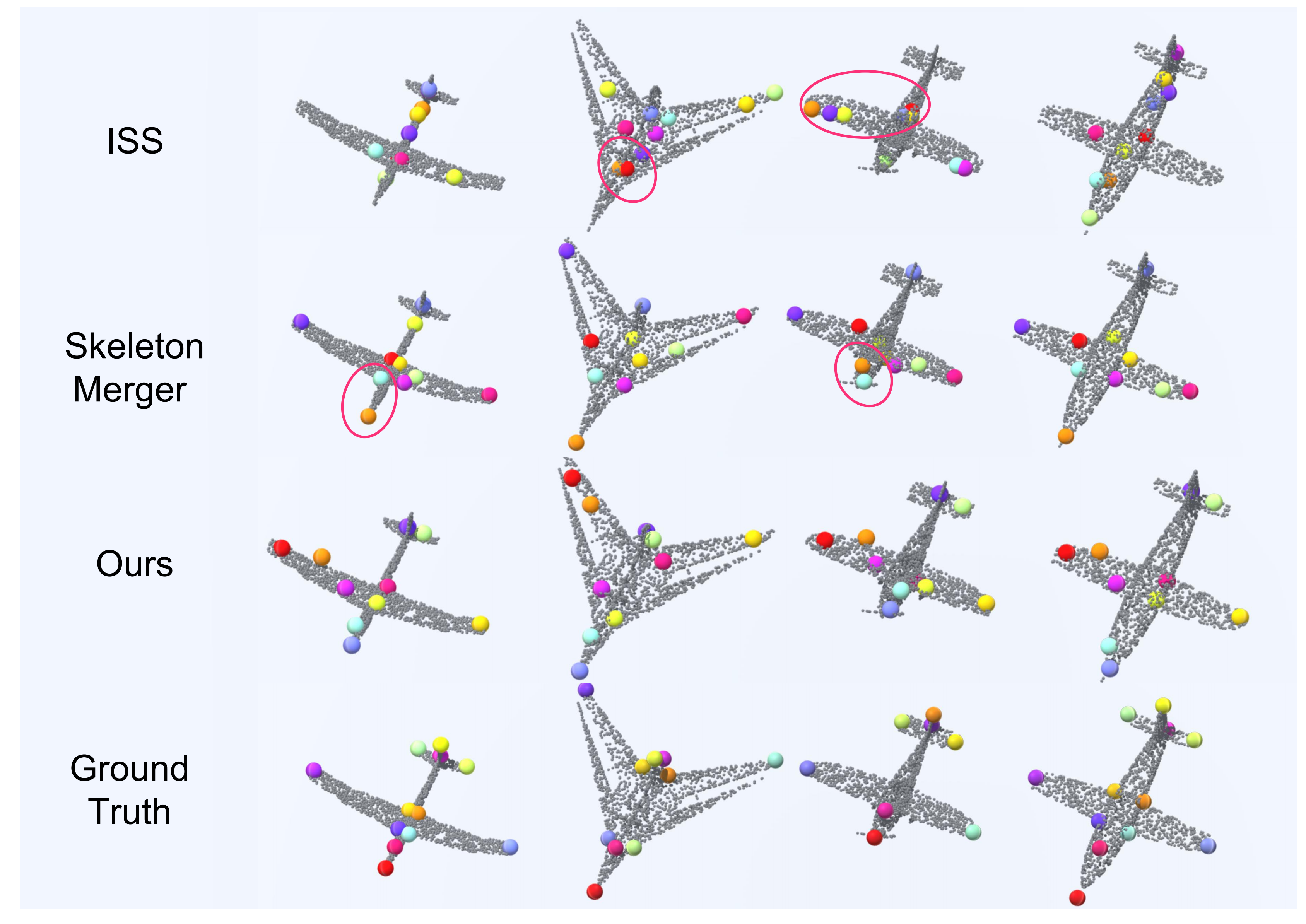}
    \caption{\textbf{Keypoints predicted by different methods.} Keypoints are rendered with different colors to show semantic consistency.}
    \label{fig:my_label}
\end{figure}

It can bee seen that our method ensures more consistent alignment of semantic keypoints, as keypoints of the same semantic label tend to be close to each other in the 3D space. Besides, the t-SNE results suggest that our encoder learns more distinctive category-level information from point clouds. Finally, we give a comparative semantic keypoint detetcion results in Fig.~\ref{fig:my_label}.

\begin{figure*}[!h]
    \centering
        \subfloat[Repeatability curve.]{
        \includegraphics[width=.39\linewidth]{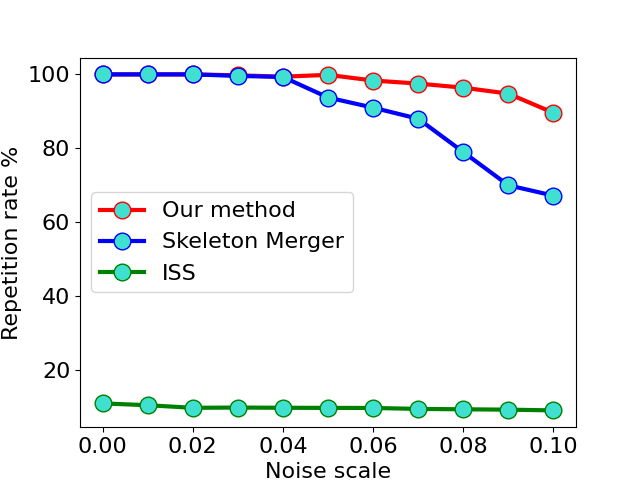}
    }
        \subfloat[Keypoint visualization under noise.]{
        \includegraphics[width=.34\linewidth]{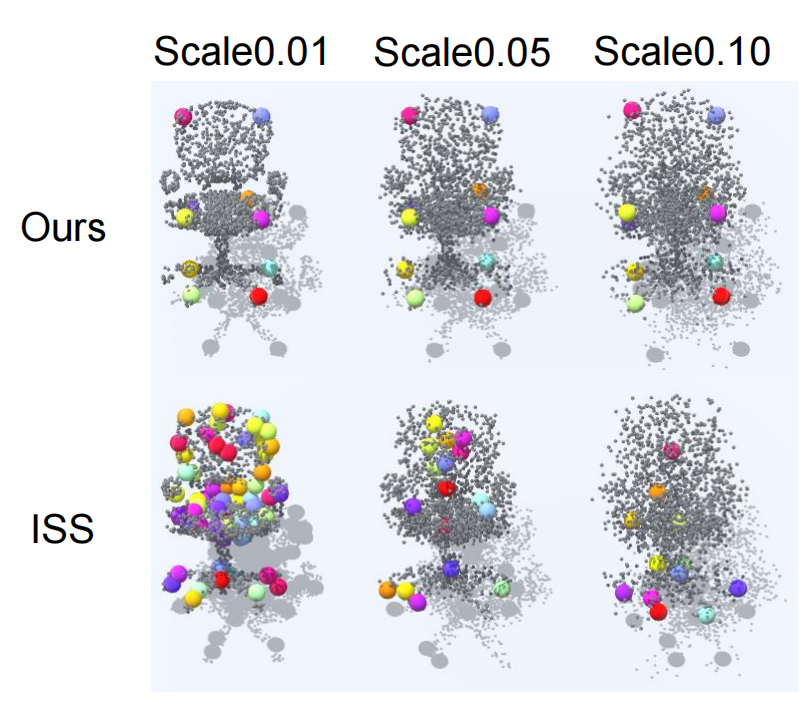}
    }
    \caption{\textbf{Robustness test.} This experiment is tested on the airplane (Fig. a) and chair (Fig. b) categories from ShapeNet\cite{shapenet}.   }
    \label{fig:noise}
\end{figure*}
\subsection{Robustness}
We test the repeatability of predicted keypoints under Gaussian noise to show the robustness of our method. Specifically, Gaussian noise with different scales are injected to the point cloud, and if the keypoint localization error on noisy point clouds are greater than a distance threshold (0.1 in this experiment), we treat the detected keypoint is not repeatable. The results in shown in Fig.~\ref{fig:noise}.

It suggests that our method holds good robustness to noise, which can be more clearly reflected by the right visualization results in Fig.~\ref{fig:noise}.

\begin{figure}[t]
    \centering
    \includegraphics[width=.7\linewidth]{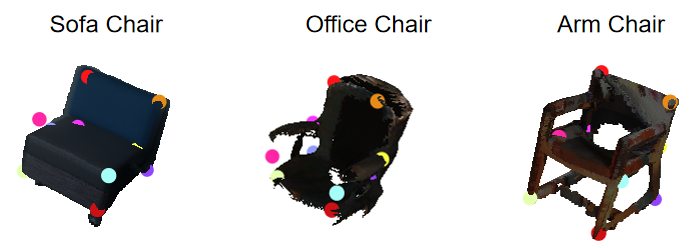}
    \caption{{\textbf{Test on a real-world scanned dataset.}} Real-world scanned ``Chairs'' are taken from the ScanNet~\cite{dai2017scannet} dataset.}
    \label{fig:realscan}
\end{figure}

We also test the generalization ability on a real-world scanned dataset~\cite{dai2017scannet}. We split chairs from the large scene in ScanNet\cite{dai2017scannet} according to the semantic label, and perform random sampling to opt 2048 points from the raw data. Our model is trained on normalized ShapeNet, and tested on the real-world scanned chairs. The result is shown in Fig.~\ref{fig:realscan}. One can see that on real-world data, the model can still predict semantic consistent points without re-training.
\subsection{Ablation Study}
To verify the effectiveness of mutual reconstruction, we compare a variation of our method without mutual reconstruction (`w/o m-rec') with the original method. The results are shown in Table~\ref{tab:my_label}.

It can be found that mutual reconstruction can significantly improve the performance as verified by both DAS and mIoU metrics. This clearly verifies the effectiveness of mutual reconstruction for unsupervised 3D semantic keypoint detection.

\begin{table}[!h]
    \centering
    \caption{Comparison of the full method and the one without mutual reconstruction.}
    \label{tab:my_label}
    \begin{tabular}{c|cccc|c}
        \toprule
         & Airplane& Chair&Car&Table&Mean \\
    \midrule
         Full method (DAS)& \textbf{81.0}& \textbf{83.1}&\textbf{74.0}&\textbf{78.5} &\textbf{79.15}\\
         w/o m-rec (DAS) & 67.2& 61.3&60.3&71.2&65.0\\
    \midrule
         Full method (mIoU)& \textbf{79.1}& \textbf{68.8}&\textbf{51.7}&54.1&\textbf{62.85}\\
         w/o m-rec (mIoU) & 77.2& 52.1&48.2&\textbf{56.1}&58.4\\
    \bottomrule
    \end{tabular}
\end{table}

\section{Conclusions}
In this paper, we proposed mutual reconstruction for 3D semantic keypoint detection. Compared with previous works, we mine \textit{category-level semantic information} from 3D point clouds from a novel mutual reconstruction view. In particular, we proposed an unsupervised Siamese network, which first encodes input point clouds into keypoint sets, and then decoding the keypoint features to achieve both self and mutual reconstructions.
In the experiments, our method delivers outstanding semantic consistency and robustness performance. Ablation study also validates the effectiveness of mutual reconstruction for unsupervised 3D semantic keypoint detection.

Though preserving global information (e.g., topology) well, the designed decoder tends to reconstruct point clouds in a skeleton-like manner, which consists limited local information. In our future work, we expect the mutual reconstruction model to be capable of detecting keypoints capturing both local and global structures.

\bibliographystyle{splncs04}
\bibliography{egbib}
\end{document}